\documentclass[10pt,journal,compsoc]{IEEEtran}

\usepackage{graphicx} 

\usepackage[hyphens]{url}
\usepackage{hyperref}
\usepackage{cleveref}
\usepackage{csquotes}

\title{Large Language Models and Games:\\A Survey and Roadmap}
\author{Roberto Gallotta$^1$~\IEEEmembership{Graduate Student Member,~IEEE}, Graham Todd$^2$~\IEEEmembership{Graduate Student Member,~IEEE}, Marvin Zammit$^1$~\IEEEmembership{Graduate Student Member,~IEEE}, Sam Earle$^2$~\IEEEmembership{Graduate Student Member,~IEEE}, Antonios Liapis$^1$~\IEEEmembership{Member,~IEEE} Julian Togelius$^2$~\IEEEmembership{Senior Member,~IEEE,} and Georgios N. Yannakakis$^1$,~\IEEEmembership{Fellow,~IEEE}\\
${}^1$: Institute of Digital Games, University of Malta, Msida, Malta\\
${}^2$: Tandon School of Engineering, New York University, New York, USA\\
roberto.gallotta@um.edu.mt, gdrtodd@nyu.edu, marvin.zammit@um.edu.mt, se2161@nyu.edu, antonios.liapis@um.edu.mt, julian@togelius.com, georgios.yannakakis@um.edu.mt}
\begin{document}
\maketitle

\begin{abstract}
Recent years have seen an explosive increase in research on large language models (LLMs), and accompanying public engagement on the topic. While starting as a niche area within natural language processing, LLMs have shown remarkable potential across a broad range of applications and domains, including games. This paper surveys the current state of the art across the various applications of LLMs \emph{in} and \emph{for} games, and identifies the different roles LLMs can take within a game. Importantly, we discuss underexplored areas and promising directions for future uses of LLMs in games and we reconcile the potential and limitations of LLMs within the games domain. As the first comprehensive survey and roadmap at the intersection of LLMs and games, we are hopeful that this paper will serve as the basis for groundbreaking research and innovation in this exciting new field.
\end{abstract}


\section{Introduction}

Five years ago, autoregressive language modeling was a somewhat niche topic within natural language processing. Training models to simply predict text based on existing text was considered of primarily theoretical interest, although it might have applications as writing support systems. This changed drastically in 2019 when the Generative Pre-trained Transformer 2 (GPT-2) model was released~\cite{radford2019language}. GPT-2 demonstrated convincingly that transformer models trained on large text corpora could not only generate surprisingly high-quality and coherent text, but also that text generation could be controlled by carefully prompting the model. While not the first autoregressive model \cite{devlin2019bert, zhang2019ernie}, GPT-2 was the first of the \enquote{large} models, and as such we use it here as cutoff mark (see also Section \ref{sec:whatis}). Subsequent developments, including larger models, instruction fine-tuning, reinforcement learning from human feedback \cite{kaufmann2023survey}, and the combination of these features in ChatGPT in late 2022, turbocharged interest in large language models (LLMs). Capabilities of LLMs were seemingly unbounded, as long as both problem and solution could be formulated as text.

LLMs are currently a very active research field. Researchers are focused on improving the capabilities of LLMs while reducing their compute and memory footprint, but also on understanding and learning to harness the capabilities of existing LLMs. Informed opinions on the ultimate capabilities of LLM technology vary widely, from the enthusiastic \cite{bubeck2023sparks} to the pessimistic \cite{mccoy2023embers,hicks2024bullshit}. Our aim is to approach the topic from somewhere in-between these two perspectives: optimistic with respect to the potential of LLMs and realistic with respect to their technical, theoretical, and ethical shortcomings.

Games, including board games and video games, serve both as a source of important benchmarks for research in Artificial Intelligence (AI) and as an important application area for AI techniques \cite{yannakakis2018ai_games}. Almost every game utilizes some kind of AI technology, and we are currently in an exploratory phase where both developers and researchers try to figure out how to best make use of recent advances in this field \cite{gwetzman2023revolution}.

In this paper, we set out to chart the impact LLMs have had on games and games research, and the impact they are likely to have in the near- to mid-term future. We survey existing work from both academia and (mostly independent) game creators that use LLMs \emph{with} and \emph{for} games. This paper does not set out to capture modern advances in LLM technology or algorithms for training LLMs. Not only do such resources exist \cite{minaee2024large}, but the breakneck speed of technical advances in this field will likely make our writeup obsolete in a year or so. Instead, we focus on work that leverages LLMs in games and propose a range of roles that the LLM can take in the broader ecosystem of games (both within the game and beyond). We lay out promising future directions for efforts to use LLMs in games, and discuss limitations (both technical and ethical) that should be addressed for a brighter future of LLM research in games.

It is important to note that this survey emerges from the top down, based on our expertise in AI and games \cite{yannakakis2018ai_games}, and extensive work on most topics covered by this paper. The focus of the paper, in Section \ref{sec:roles}, is built from our own typology and supported where possible by academic and non-academic work. While a bottom-up approach via e.g. keyword search through general paper repositories is valuable \cite{garciasanchez2019study,liapis2020tenyears,jordanous2016corpus}, the process would lead to a very different type of paper. Indicatively, this approach was followed by Yang \textit{et al.} \cite{daijin2024gpt} who investigated current uses of GPT models in video games by searching articles in ACM, IEEE Xplore, Springer, and AAAI with keywords \enquote{game} and \enquote{GPT}. Instead, we have attempted to conduct a comprehensive manual review of all recent proceedings from the major conferences in AI and games\footnote{IEEE Conference on Games, Foundations of Digital Games conference, Artificial Intelligence and Interactive Digital Entertainment conference.}, and the IEEE Transactions on Games for work relevant to the themes of this paper.

\section{A Note on Terminology}\label{sec:whatis}

This paper concerns the intersection between games (board games, video games, or other), and large language models. But what exactly is an LLM?

Broadly speaking, an LLM is a model that is trained on text in order to be able to reproduce text in response to other text. But this definition is overly broad, as it would include Shannon's original $n$-gram models from 1946~\cite{shannon1948mathematical}, rudimentary recurrent neural networks from the early 1990s~\cite{elman1990finding}, and the Tegic T9 text prediction system that would help you write text messages on your Nokia 3210.

What distinguishes LLMs from other text generative models is mainly that they are \emph{large}. But which model size is considered large enough? In 2019 emerging models such as BERT \cite{devlin2019bert} and ERNIE \cite{zhang2019ernie} showcased significant advances in language modeling, but LLMs became a well-recognized term with the introduction of OpenAI's GPT-2 ~\cite{radford2019language}, whose various versions have between 117 millions and 1.5 billions of parameters. Because of the association between the term LLM and the GPT-class of models, we will use the size of GPT-2 as a soft cutoff on the type of models we consider LLMs; we are concerned with models of few hundred million parameters or more. Each subsequent iteration of the GPT family features an increased number of parameters, and larger and more diverse training corpora.

Another distinct trait of LLMs is their architecture. While language models could in principle be based on various architectures, including Long Short-Term Memory (LSTM) networks \cite{Hochreiter1997lstm}, the current LLM landscape is dominated by variants of the transformer architecture, a type of neural network introduced in 2017 \cite{vaswani2017attention}. This model became very influential because of what was perceived as a quantum leap in output quality compared to previous models. In this survey we rely primarily on LLMs employing this architectural basis.

The last feature of LLMs we consider is their versatility across a wide range of tasks with minimal or no fine-tuning or retraining. This capability represents a significant shift; since the release of GPT-3.5, LLMs have evolved from primarily autoregressive predictive text models to pre-trained, general-purpose conversational models.

It is important to note that LLMs are by no means limited to the GPT family of models. There is by now a large variety of LLMs of varying size and capabilities, including open-source models such as Mistral \cite{jiang2023mistral} and the Llama \cite{touvron2023llama} family, which can be fine-tuned, run locally, and even be embedded in games' runtimes.

One could also argue that the definition is somewhat narrow, as many modern LLMs are multimodal models, meaning that they can take as input and/or produce as output modalities other than text. In particular, many modern LLMs can process and produce images. This is often achieved through combining the core transformer network with a visual encoder network for input and a latent diffusion model for output. Examples include GPT-4V \cite{yang2023dawn} and the open-source Llava \cite{liu2023visual}. In this paper, we consider large multimodal models (LMMs) \cite{yang2023dawn} as long as they retain their ability to both consume and produce text.

This survey will not concern itself with AI and machine learning techniques that are not LLMs as defined above. In particular, we will not be covering the large literature on game playing and content generation using machine learning methods \cite{summerville2018pcgml} that does not use textual input and output. We will, however, occasionally mention some of that work where relevant, in particular to help provide historical context.

\section{Roles of LLMs in Games}\label{sec:roles}

Past attempts at a typology for AI in games focused on three roles the AI can take in a game: to play a game, to design a game, or to model the (human) players \cite{yannakakis2018ai_games}. LLMs are typically presented as conversational agents, which often invites the public to give them anthropomorphic qualities \textemdash such as reasoning and creativity. We follow these trends when considering the roles an LLM can be called to play within the game or within the game development process. An LLM can operate within the game as a player (replacing a human player while imitating their goals), as a non-player character such as an enemy or interlocutor, as an assistant providing hints or handling menial tasks for a human player, as a Game Master controlling the flow of the game, or hidden within the games' ruleset (controlling a mechanic of the game). There are however other roles an LLM can play outside of the game's runtime, such as a designer for the game (replacing or assisting a human designer) or as analyst of the gameplay data of the playerbase. Finally, the LLM can interface with a player or an audience in different ways, acting as a commentator of an ongoing play session (during runtime) or a reteller of past game events in some narrative form (outside runtime). Some of these roles (autonomous player, autonomous designer) are prominent in the broader AI and games research \cite{yannakakis2018ai_games} and LLM research has targeted them extensively, while some of the other roles have been toyed with in exploratory research. The following sections present the roles themselves, surveying research undertaken for each role, while we identify gaps and opportunities for future research in Section \ref{sec:roadmap}.

\subsection{Player}\label{sec:player}

How can an LLM play a game? Fundamentally, LLM players require some transformation from their typical output space (i.e. sequences of tokens) into the input space of the game. In addition, aspects of the game and its current state must be provided to the LLM in some form in order for it to play at a reasonable level. Depending on the game itself, these mappings might be intuitive or complex. We identify three general classes of games to which LLM players are well suited: (a) games where states and actions can be compactly represented as sequences of abstract tokens, (b) games where the main input and output modalities are natural language, and (c) games for which external programs can control player actions via an API. 

The first class of games mostly includes turn-based board games (e.g. \textit{Chess}), since the discrete set of board positions and moves is more easily transformed into a compact representation (e.g. Portable Game Notation \cite{pgn}) than, for instance, a first-person shooter. By tokenizing sequences of moves taken from a game database, the problem of action selection can then be mapped to the standard autoregressive learning objective on which LLMs are trained \textemdash predicting the next move given the context of those that preceded it. \emph{Chess} \cite{noever2020chess, stockl2021watching, toshniwal2022chess}, \textit{Go} \cite{ciolino2020go}, and \textit{Othello} \cite{li2022emergent} have all been used as testbeds for LLM players in this way. This approach allows even more complex game states to be reasoned upon by an LLM player. In Bateni and Whitehead's work \cite{bateni2024language}, for example, the LLM plays the popular video game \emph{Slay the Spire} (Mega Crit, 2017), understanding synergies between cards based solely on their description and adapting to changes in gameplay rules. However, board games are not the only kind of game that can be represented as token sequences: the generalist {GATO} \cite{reed2022generalist} agent can play a variety of Atari games at human or near-human levels by processing visual inputs as sequences of pixel values in raster order. Pixel values are interleaved with separator tokens and previous actions, allowing the model to accurately predict the appropriate game action in a dataset of human play traces. It is possible that continued improvement in transformer models that capture both spatial and visual dynamics \cite{xu2020spatial, chang2022maskgit} could allow for a similar approach to scale to even more complex games. However, such approaches require large datasets of gameplay videos that may be comparatively more difficult to collect. In addition, we note that reliance on human gameplay traces as the basis for learning may make it more difficult for an LLM player to reach super-human performance without leaps in terms of reasoning and generalization (see Section \ref{sec:limitations}).

The second class of games most obviously includes text adventure games such as \textit{Zork} (Infocom, 1977), where game states are presented as natural language descriptions and the game is already equipped with a parser to handle natural language responses. This means that LLMs can be queried for game actions in a way that still leverages their large-scale pre-training on natural language text. The earliest application of LLMs to these kinds of text games is CALM \cite{yao2020keep}, a GPT-2 system finetuned on a dataset of human gameplay transcripts collected from a variety of text adventure games. The model is trained to predict the natural language string provided by human players given the context of previous states, actions and information about the avatar (e.g. their inventory). To actually play a game, the trained language model generates multiple candidate actions and deep reinforcement learning (RL) is used to optimize a policy that selects actions from among the candidates. At the time of its publication, this RL component was necessary because the LLM alone was not capable of generalizing well to unseen games or situations \cite{yao2020keep}. However, a more recent investigation of ChatGPT as a \textit{Zork} player has indicated that LLM performance is improving \cite{tsai2023can}. In a preliminary experiment, Yao \textit{et al.} \cite{yao2020keep} show that the performance of ChatGPT can approach that of existing algorithms for text game playing, as long as a human interlocutor remains in the loop to assist the model (e.g. by reminding it of actions it has already tried). However, there is obviously much room for improvement in directly applying LLMs to text games in this way. Additionally, the ability for LLMs to play entirely novel, niche, or unseen text games (especially important given the likelihood that such systems encounter walkthroughs or playtraces of popular text games during their training) remains largely unexplored. 

In a similar vein, inductive biases from large language models can be applied to help guide the policies of agents trained with other methods. For instance, the GALAD system \cite{ammanabrolu2022aligning} uses a pre-trained LLM to guide an agent towards morally acceptable actions in text games from the Jericho suite \cite{hausknecht2020interactive}, while the MOTIF system \cite{klissarov2023motif} learns an intrinsic reward function for \textit{NetHack} NetHack, 1987) by mining preferences between game states from an LLM.

Text adventure games are not the only cases where natural language input and output are used for playing: many board games operate via player negotiation. \emph{CICERO} \cite{meta2022human} leveraged LLMs for playing the deal-making and subterfuge game \emph{Diplomacy} (Avalon Hill Games, 1976). {CICERO} builds from a pre-trained LLM and is fine-tuned on a large corpus of \emph{Diplomacy} transcripts. Throughout the game, samples from the model are sent to other players and the various dialogue transcripts are collected to condition the potential action. {CICERO} is further trained to condition its outputs on specific game intents (inferred from the transcripts and added as additional context during training). In order to select an action, {CICERO} uses a \enquote{strategic reasoning module} that predicts the actions of other players, using a value and policy function learned from self-play. \textit{Diplomacy} is an interesting game in part because the action space is split between natural language utterances and a more standard set of moves on a discrete game board, and {CICERO} demonstrates how an LLM can be integrated as part of a larger system for high-level play.

Finally, we consider games for which a robust API exists. This is less of a \textit{kind} of game in the sense of its style or mechanics, and more a fact about its popularity or its ease of implementation. An API is an important attribute because it allows LLMs to act as players not by directly generating actions, but by producing \textit{programs} that act as \textit{policies}. Improvements in the code generation abilities of LLMs have allowed them to write small programs that can produce actions given game states without further intervention from the model. For instance, the {VOYAGER} system \cite{wang2023voyager} leverages the code generation abilities of GPT-4 to play \textit{Minecraft} (Mojang Studios, 2011) by interacting with the popular Mineflayer API. Using a sophisticated chain of prompts, {VOYAGER} generates blocks of code that leverage calls to the API in order to execute high-level \enquote{skills} (e.g. \enquote{Attack nearest zombie}) that are automatically converted into low-level game inputs (e.g. mouse movements and key presses). GPT-4 is also used as a high-level goal generator and planner, which in turn informs the code generation. This approach proved to be very successful, with {VOYAGER} being the first automated system to complete a variety of in-game \emph{Minecraft} challenges. The results are impressive and indicate that generating action-producing programs may be a more efficient way to leverage latent LLM knowledge than direct action sampling. However, {VOYAGER} does benefit substantially from the availability of a robust API and vast amounts of internet discussions around \emph{Minecraft}. As with the analysis of ChatGPT on \emph{Zork}, the ability of this approach to generalize to less popular or entirely unseen games remains to be seen. 

\subsection{Non-Player Characters}\label{sec:npcs}

\begin{figure}
    \centering
    \includegraphics[width=0.98\columnwidth]{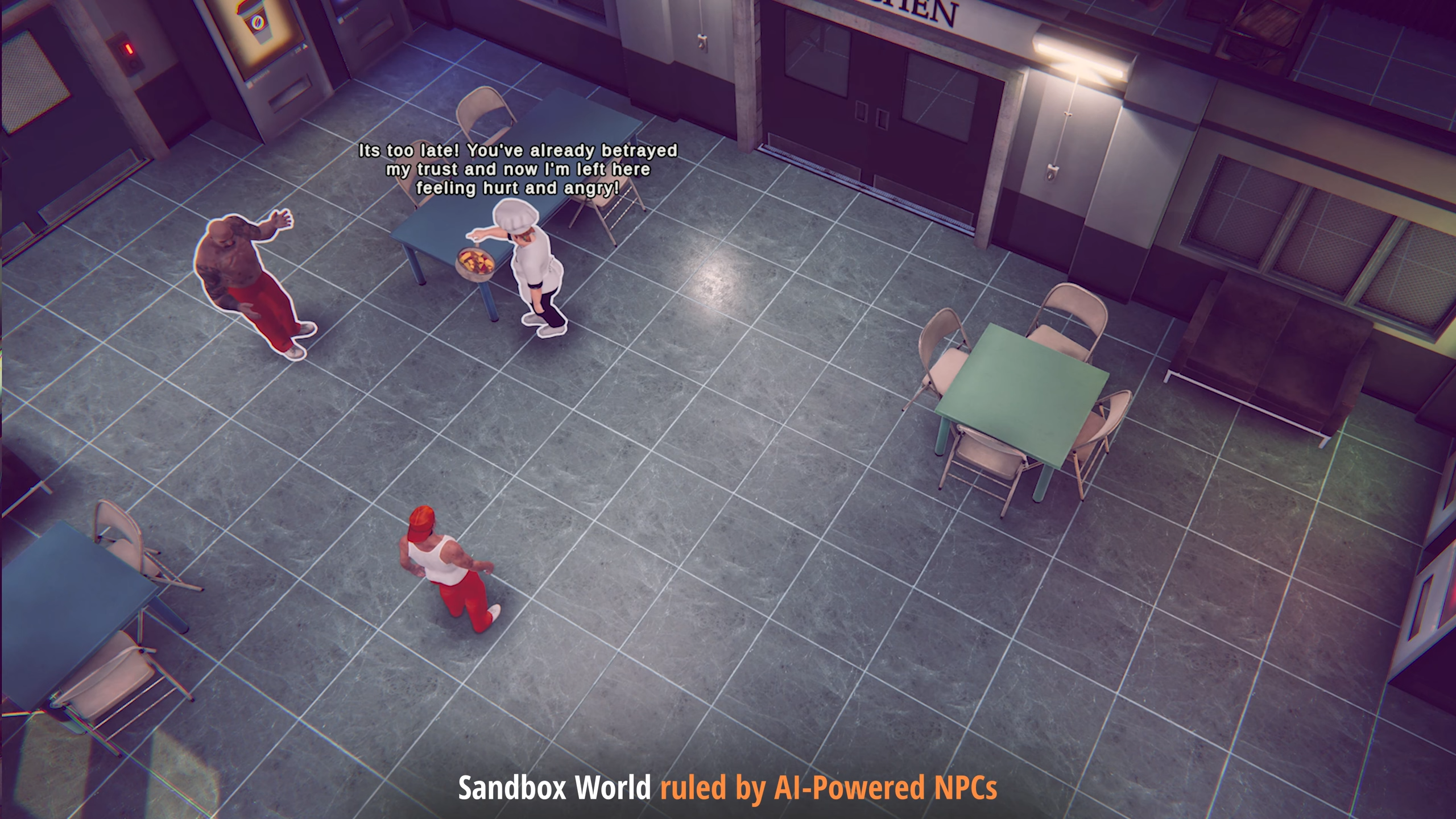}
    \caption{Screenshot of the promotional video for \textit{AI people}, where a player can interact via their avatar's chat with other NPCs and watch how their text has consequences between NPC relationships, in this case. Used with permission from \cite{goodai2023blog}.}
    \label{fig:aipeople}
\end{figure}

Non-player characters (NPCs) are agents which exist in virtual game worlds but whose actions are not directly controlled by the players. NPCs exist to enrich the player's experience and deepen immersion by adding to the world's ambiance and making it more believable \cite{Uludagli2023npc_decision_making}. NPCs may serve as pets, allies, enemies\footnote{We use enemies and allies here for in-game agents with different skills and ways to affect the world than e.g. game playing AI such as opponents in \emph{Chess}.}, merchants, quest givers, or bystanders. Therefore, they have different agency even from AI-controlled players, and their goal is never to win. This makes designing AI for NPCs interesting \cite{yannakakis2018ai_games}, and LLMs are uniquely advantaged in this task since they can \enquote{understand} gameworld settings and adapt their responses accordingly. It has been shown that LLMs are able to role-play through different scenarios \cite{Shanahan2023role_play_LLMs,gao2023chemical}, thereby highlighting their potential to provide a more flexible and apt tool to emulate human behavior. We identify two ways in which LLMs can control NPCs: (a) through their \emph{dialogue}, and (b) through their \emph{behavior}. Behavior relates to in-game action selection, discussed in Section \ref{sec:player}; however, we note that the heuristics and goals of such behavior is different than an AI player trying to win the game.

LLMs are naturally suited for natural language conversation, and as NPC dialogue systems they can generate dynamic and contextually appropriate responses based on player input \cite{park2023simulacra,kreminski2023toward,muller2023chatter,obrien2024neonpc}. This makes interactions with NPCs more engaging and realistic, reduces repetitive discourse and provides a more explorative experience within the game \cite{Warpefelt2107believability}. LLMs can engage the players in the gameworld's narrative as \emph{foreground} NPCs, \emph{background} NPCs, or \emph{narrator} NPCs. We discuss narrator LLMs as commentators in Section \ref{sec:reteller} whereas we cover the other two NPC types here. Foreground NPCs form part of the overarching narrative of the game, or one of its sub-narratives. They may be enemies, allies, information-givers, quest-givers, or item-providers. Their dialogue is heavily constrained by the scope of the narrative, their role within it, and the player actions. Foreground NPCs' text generation process via LLMs must consider the overall context of the game and the interaction with the player, and keep track of events transpiring in the playthrough. This raises concerns regarding the memory capacity of LLMs, as well as the impact of possible hallucinations, i.e. plausible but false statements \cite{duan_2023_shifting}. We revisit these limitations in Section \ref{sec:limitations}.
The purpose of background NPCs is to make the environment more believable and act independently of the players \cite{Warpefelt2107believability}. Such NPCs' presence is purely decorative and their dialogue is essentially small talk. Thus their dialogue generation is less constrained, perhaps bound only by the identity of the speaker and their background. That said, their believability hinges on their ability to maintain the illusion that they have their own agency in the world and can interact with it \cite{Mehta2021LLMsNPCsurvey}. Park \textit{et al.} \cite{park2023simulacra} explored NPC interactions within limited environments, where a number of LLM-based agents simulated social interactions in a sandbox environment. Within the constraints of their environment and the social affordances, the agents behaved in a believable manner, following their goals, planning for new ones, and even recalling past events as they interacted.

Other studies have shown that multiple LLM-based agents are able to follow game rules and engage in game playing \cite{akata2023playing,xu2023exploring}, with different models consistently exhibiting their own aptitudes and weaknesses when applied to specific roles. This ability to interact within constraints is useful to instill believable behaviors in foreground and background NPCs, grounding their actions and dialogue within the rules of the game environment. Some work has focused more on the conversational and story-writing abilities of LLMs, such as the creation of dialogue between multiple characters, each having their unique personality, whilst following a consistent plot. One such example is the use of LLMs to generate a \emph{South Park} (Comedy Central, 1997) episode \cite{fable2023showrunner} with multiple characters within a well-known setting. There are limitations to this approach, primarily that LLMs perform something like a theatrical improvisation, rather than acting as an actor studying a part \cite{Shanahan2023role_play_LLMs}. Through this unconstrained process the LLM is prone to hallucinations which do not fit the desired scenario. This volatility can be mitigated by providing the LLM not only with the conversation history but also with the current state of the environment, such as the items within it and their affordances, as well as the other characters and their corresponding actions. Urbanek \textit{et al.} \cite{urbanek2019learning} used a configurable multi-user dungeon text-based environment in a fantasy setting to allow for both human and language model-based players. The latter were able to use an updated game state, including descriptions of local environments, objects, and characters, to take better actions and engage in coherent dialogue. This approach may also be extended to other scenarios or to cover the use of LLMs as active or interactive narrators. Ubisoft showcased LLM-based NPCs in their Neo NPC demo \cite{obrien2024neonpc}, where the player can freely converse with the in-game characters. Each NPC was given a carefully hand-crafted persona, and all NPCs could respond within the constraints of the game narrative and their prescribed personality whilst generating realistic responses. The players were able to engage in game-specific activities with these NPCs, such as planning a heist, or even attempt an unrelated course of dialogue altogether. Care was taken to minimize toxicity and social biases inherent in the LLMs through the prompt defining each NPC, which also had a bearing on how the latter would react to any player's offensive or unruly discourse.

\subsection{Player Assistant}\label{sec:player_assistant}

A somewhat less explored role for LLMs in games is that of a player assistant: an interactive agent intended to enrich or guide the player experience in some way. This could be in the form of a sequence of tutorial-style tips, a character that does not causally interact with the game world at all, or an agent able to interact within the game environment at a similar level as the player. Existing games make use of player assistants in different ways. For example, in \emph{The Sims} (Electronic Arts, 2000) a disembodied assistant provides tips specific to the game context via dialogue boxes. \emph{Civilization VI} (Firaxis Games, 2016) uses different assistants (with different portraits) to suggest the best build option according to their idiosyncratic heuristic, alleviating some decision-making from the player. In management games, AI may automate menial tasks such as assigning jobs to a planet's population in \emph{Stellaris} (Paradox Interactive, 2016); this assistance reduces cognitive load from the player, but the player can always micro-manage this task if they wish.

LLMs are appealing as player assistants given their expressive and conversational capacities. An LLM-based player assistant could plausibly choose an action to suggest to the player, and---more importantly---form the explanation for this suggestion as a natural language utterance delivered by a disembodied or embodied agent. This utterance could even be accompanied by a corresponding sentiment, and manifested through the assistant's body stance, gestures and facial expression (in the case of an embodied assistant). The choice of action to suggest to the player could be based on either LLM-powered or heuristic-based methods for finding the best policy or action given the current game context (see Section \ref{sec:player}). While not intended as a player assistant, this type of exposition is realized by embodied agents in \emph{AI people} \cite{goodai2023blog} (see Section \ref{sec:mechanic}). Similarly, LLMs may assist the player by undertaking some minor tasks in the game via a tailored smaller role as ``player'' within that smaller task description (see Section \ref{sec:player}). The LLM could extrapolate the policy for such a minor task through conversation with the player, parsing the natural language chat similar to \cite{hu2024minecraft}. 

A special case of player assistant comes from an ``inner voice'', uttered by the player's own avatar. Hints given in the player avatar's own voice are a trope in classic point-and-click adventure games, such as Guybrush Threepwood commenting ``A rubber chicken with a pulley in the middle{\ldots}What possible use could that have?'' when picking up the item in \emph{The Secret of Monkey Island} (Lucasfilm Games, 1990). This specific concept of \enquote{inner voice} was explored as an application of LLMs in the work of Rist \cite{rist2024using}, although the LLM's freedom was limited. Using a hand-crafted game environment with pre-scripted events and location-based triggers meant to occur when the player would either interact with the world or observe something, Rist authored where the LLM is prompted to generate short comments in different styles (e.g. in neutral or sarcastic tone) when certain predefined in-game events are met. These comments are generated based on text descriptions of what the user sees and can do in that moment. The LLM still requires human-authored knowledge (e.g. \textit{where} the commentary occurs, \textit{what} the hints are, and \textit{why} in terms of designer goals), and is thus not a fully-autonomous player assistant.

Despite the work of Rist \cite{rist2024using}, which focused more on immersion than assistance, the potential of LLM-powered player assistants is not explored in current research. We highlight the potential of this application in Section \ref{sec:roadmap}.

\subsection{Commentator/Reteller}\label{sec:reteller}

LLMs are also ideally suited as commentators or retellers. Here, we identify these roles as an agent that produces and narrates a sequence of events, for the benefit of either human players or spectators. Such an agent may consider only in-game events and in-game context, acting as an in-game entity such as a sports commentator in \emph{FIFA} (EA Sports, 1993) or also consider out-of-game events and context such as the player (their actions, strategies, motivations, etc.). The \textit{reteller} \cite{eladhari2018retellings} exclusively narrates past events \textemdash often grouped into a concise \enquote{chunk} such as a game session (i.e. based on out-of-game context) or a quest (i.e. based only on in-game context). The \textit{commentator} may be narrating current, ongoing events which have not been concluded, similar to a streamer concurrently discussing their current actions (including out-of-game context) or a sportscaster in an in-progress sports game such as \emph{FIFA}. 

The vision of automated \enquote{let's play}-style commentary generation is not new. It was proposed by Guzdial, Shah and Riedl \textit{et al.} \cite{guzdial2018towards} and implemented via classical machine learning methods, with limited success. Ishigaki \textit{et al.} \cite{ishigaki2021generating} trained an LSTM with text, vision and game state input to generate characters for a commentary script in the racing game \textit{Assetto Corsa} (Kunos Simulazioni, 2013). Results of this approach featured repetitive and context-irrelevant generated text. LSTMs were also used by Li, Gandhi and Harrison \cite{li2019end} to generate text, at a character level, for \emph{Getting Over It With Bennett Foddy} (Bennett Foddy, 2017), a challenging side-scrolling climbing game.

LLMs for commentary were also explored by Renella and Eger \cite{renella2023towards}, who argue that LLMs could assist game streamers (e.g. on Twitch) while the streamer multitasks gameplay with audience interaction. The authors developed a pipeline for automatically commenting upon \emph{League of Legends} (Riot Games, 2009) games. They took a multi-phased approach, training a model on hand-annotated data to recognize key events, then prompting ChatGPT to generate zero-shot commentary on these events in the style of a particular (known) fictional character, and finally sending the generated text through the FakeYou\footnote{\url{https://fakeyou.com/}} API to be voiced in the timbre of this same character. For example, once the event detection model has identified an enemy double kill in a particular frame, ChatGPT responds in the style of Rick Sanchez from \emph{Rick and Morty} (Cartoon Network, 2013): \enquote{What the heck?! That enemy team just got a double kill! I can’t believe it! They must be pretty good! I better watch out for them!} An additional loop buffers detected events \textemdash for example, delaying commentary on a double kill in case it should escalate into a triple kill, or prioritizing among a quick barrage of events \textemdash and prompts ChatGPT to generate random fillers, such as thanking (fictional) new subscribers.

Despite the existence of the aforementioned studies, research on LLMs as game commentators remains rather limited. The appeal is obvious: simulation-based games of emergent narrative already generate rich narrative histories, and are remixed by human players to produce secondary content that is often popular in its own right. In principle, LLMs could be used to generate more succinct retellings or highlight reels of these game events. Prompting current LLMs for stories, without any further specification of style or substance, tends to produce output that feels generic. Past events recorded in simulation games could ultimately provide specificity and narrative coherence to these outputs. Exploring more concepts beyond automating streamer commentary, such as assisting streamers via LLM commentary of the audiences' reactions rather than the in-game actions, remains unexplored. We revisit this along other future applications in Section \ref{sec:roadmap}.

\subsection{Analyst}\label{sec:analyst}

Another role for LLMs is that of a data analyst. Here, we take this role to primarily analyze player experience and behavior, rather than the broader big-data job title for which some LLMs are naturally suited for \cite{jaimovitch2023can, cheng2023gpt,maddigan2023chat2vis}. While \textit{player modeling} is an important aspect of AI research \cite{yannakakis2018ai_games} and game development practice \cite{seifelnasr2004analytics}, LLMs have not received much attention for this purpose. However, the ability of LLMs to make sense of structured data such as code \cite{cheng2023binding, chen2023exploring} can be a boon for this type of work.

So far, language models (both large and small) have been used for clustering player behaviors. Player2vec \cite{wang2024player2vec} trained the Longformer transformer architecture \cite{beltagy2020longformer}, with up to 121 million parameters, on a corpus of game events stored in JSON format which captured player interactions with a casual mobile game. Using dimensionality reduction methods on the latent vectors of the last layer of a pre-trained transformer, the authors discovered eight clusters with player traits that could be useful for market research, such as ``lean-in casual economy aware'' \cite{wang2024player2vec}. In such work, the strength of LLMs is their ability to process structured or unstructured data without the need for extensive data pre-processing. However, as with previous experiments in bottom-up clustering of players \cite{canossa2009tombraider,canossa2013dig}, the resulting clusters need to be interpreted by expert game analysts (or the game's designers) to derive some meaningful player types. In this form, current research points to LLMs acting as \textit{analyst assistants} rather than independent analysts; we revisit future directions in this vein in Section \ref{sec:roadmap}.

Representing game logs as text interpreted by an LLM need not be the end-goal for analysis. The LLM representations of game logs can also be used to find common patterns that can be used for other purposes such as gameplay (or gameplay footage) similarity. Indicatively, Rasajski \textit{et al.} \cite{rasajski2024behave} used LLMs on recorded gameplay action logs to establish action similarity. This similarity was in turn applied to train video encodings on gameplay footage of these recorded actions, in order to align the visual latent vectors to the action logs. While the work focused on the computer vision task of representation learning for gameplay footage, better (learned) representations of gameplay pixels would be more generalizable across games in downstream tasks. Such latent representations would be useful for general action recognition or general affect modeling \cite{togelius2016generalgeneral}; we expand on these downstream tasks in Section \ref{sec:roadmap}.

\subsection{Game Master}\label{sec:gamemaster}

A Game Master (GM) in tabletop role-playing games (TTRPGs) is the person who creates the plot of a game, its characters, and narrative. GMs wear many hats during the course of the game session \cite{tychsen2005gamemaster}; they prepare and adapt stories before sessions, guide gameplay during, and follow up with players afterward \cite{liapis2023playerexperience}. Digital games have mostly pre-scripted stories or level progressions and their players have a restricted range of affordances, compared to TTRPG players whose actions are only limited by their imagination. Similarly, the story told around the table can take any direction. Since human GMs mostly communicate about the gameworld, story, game state and action resolutions via natural language (although props such as maps, miniatures, hand-outs are also common), the potential of LLMs as a GM is often mentioned both in research circles and TTRPG discussion boards. LLMs as GMs also open the potential for solo play, while a TTRPG requires at least one player and a human GM.

One of the first notable text adventures managed by a fine-tuned version of GPT-2 is \textit{AI Dungeon} \cite{hua_playing_2020}. It is an online\footnote{\url{https://play.aidungeon.com/}} interactive chat-based storytelling application where the player takes actions through language input alone. The LLM continues the story based on the player's input, in the fashion of a human GM. The game has evolved since its creation to make use of more recent LLM models, which the player can choose from before starting a play session. Different gameworld settings are also offered, and players are also able to share the stories they create. Similar games have emerged online ever since\footnote{\url{https://koboldai.net}, \url{https://www.hiddendoor.co}}, and a freely available code repository, \emph{Kobold AI Client}\footnote{\url{https://github.com/KoboldAI/KoboldAI-Client}}, allows a local or remote installation of a client for such LLM-run games. Some of these games also use Stable Diffusion text-to-image models \cite{Rombach_2022_StableDiff} to generate visuals accompanying different parts of the narrative. More recent work \cite{you2024dungeons} investigates how different characterizations of the GM and their way of presenting events to the players impacts the overall experience in TTRPGs. This study presented an online interface for a custom game based on a \emph{Dungeons \& Dragons} (TSR, 1974) fantasy setting. LLMs were ascribed different GM roles through a number of prompts, and human players participated in 45-minute games (split into 3 sub-sessions), with an overall positive response to the automated GM.

In lieu of replacing a human GM, LLMs have also been employed as GM assistants. CALYPSO \cite{Zhu2023Calypso} is a set of tools running on a Discord server which the GM can query either to generate random encounters, brainstorm ideas, or alternatively chat with a fictional character in a \emph{Dungeons \& Dragons} setting. CALYPSO highlights that hallucinations of GPT-3 can have both positive effects when it generates plausible details not included in descriptions published in the original game manual (e.g. the shapes of creatures' eyes) and also negative effects when the created details are outright incorrect (e.g. describing the wings of a canonically wingless creature). In addition, the model's preconditioning to avoid racial bias was found to occasionally prevent it from generating racial details of fantasy creatures in the game.
Other work used smaller GPT models to improvise in-game conversations \cite{Kelly2023TTRPGsupport} by monitoring and transcribing verbal exchanges between the GM and the players, and attempting to generate appropriate responses. This example was integrated into \textit{Shoelace} \cite{Acharia2023shoelace}, which is itself a GM assisting tool helping with content lookup by creating a node-based plan of the game narrative and encounters. The versatility of LLMs given their ability to rapidly process text input paves the way for their integration into the multitude of existing tools and aids for human GMs.

\begin{figure}[t]
    \centering
    \includegraphics[width=0.98\columnwidth]{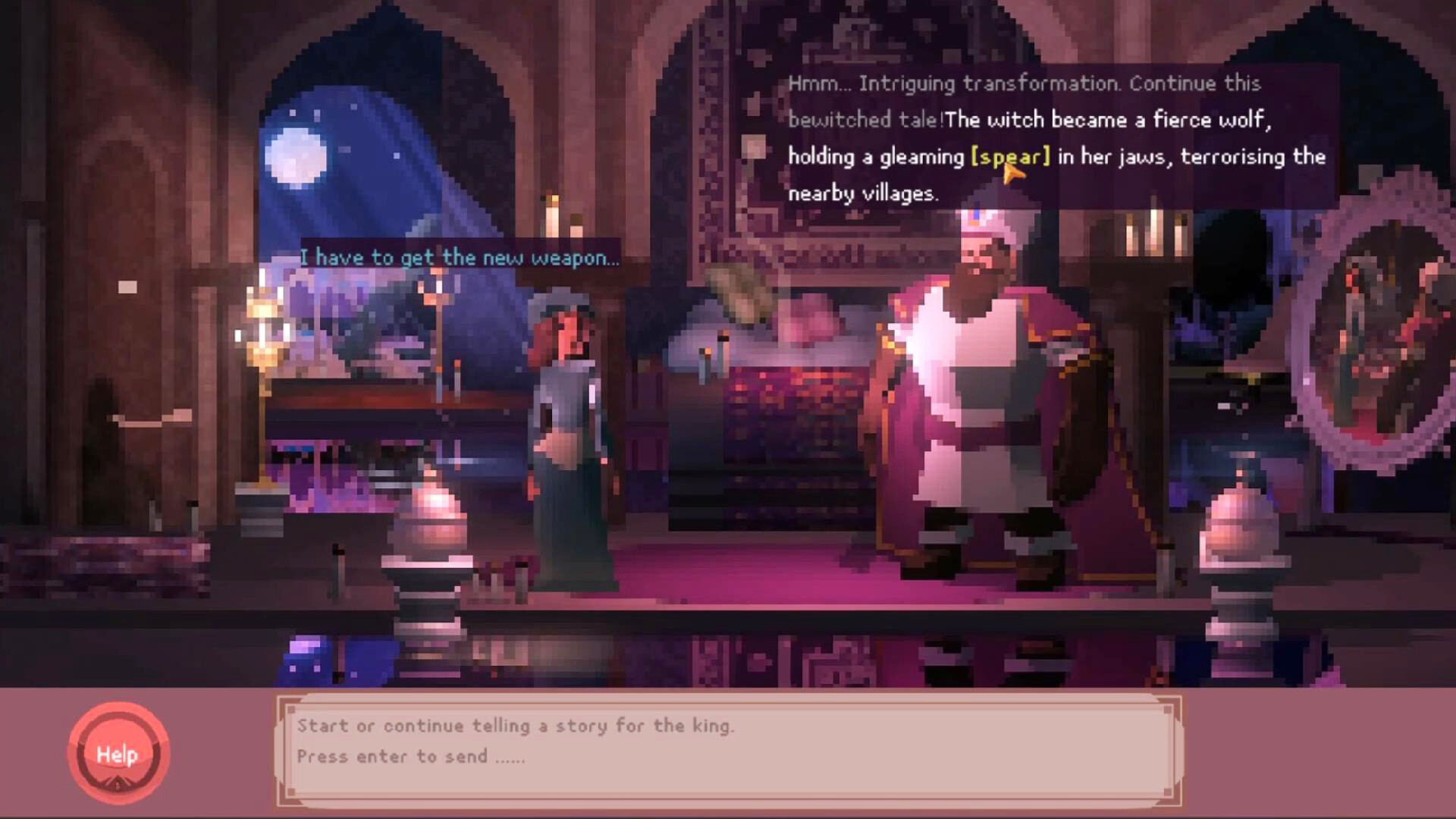}
    \caption{In \textit{1001 Nights}~\cite{sun2023language}, the player uses free-form text to trick the king (role-played by an LLM) into uttering the name of a particular weapon (which will then materialize, allowing the player to defeat him). Image used with permission.}
    \label{fig:1001_nights}
\end{figure}

\subsection{Game Mechanic}\label{sec:mechanic}

Games can also be built around a specific mechanic that relies on LLMs, similar to the AI-based game design patterns identified by Treanor \textit{et al.} \cite{treanor2015aigamedesign}.
An obvious mechanic revolves around the social interactions facilitated by LLM-powered conversational NPCs. In this vein, the \textit{Generative Agents} project \cite{park2023simulacra} has employed LLMs to populate a virtual village with 25 characters, enabling them to communicate and engage in social behavior within a sandbox environment. Players were able to interact with these agents via text. The environment state and actions of each agent were stored in a language-based format and summarized in order to retain knowledge for each agent when prompting for its actions. This led to emerging believable social interactions, such as the agents spontaneously inviting other agents to a party which one of them was organizing. Similarly, GoodAI are developing the \textit{AI people} video game which operates as a sandbox simulation where LLM-powered NPCs \enquote{interact with each other and their environment, forming relationships and displaying emotions} \cite{goodai2023blog}. The player can interact with the agents via natural language chat, triggering reactions and potentially disrupting the relationship between NPCs (see Fig.~\ref{fig:aipeople}).

Natural language interactions form a natural pool of mechanics to build games around, such as gamifying users' attempts at jailbreaking LLMs \cite{liu2023jailbreaking}. The game \textit{1001 nights}, depicted in \Cref{fig:1001_nights}, exemplifies this by having an LLM co-create a story from human prompts, where the player's objective is to try and steer the story to include specific keywords in order for the main character, Scheherazade, to turn these into tangible items in aid of her escape \cite{sun2023language}. Similarly, \emph{Gandalf}\footnote{\url{https://gandalf.lakera.ai/}} challenges the player to trick an LLM into revealing a password. The game increases the difficulty of the task as levels progress by adjusting the prompt specifications, such as forcing the LLM to re-examine its generated response to ensure it does not include the password.

\begin{figure}[t]
    \centering
    \includegraphics[width=\columnwidth]{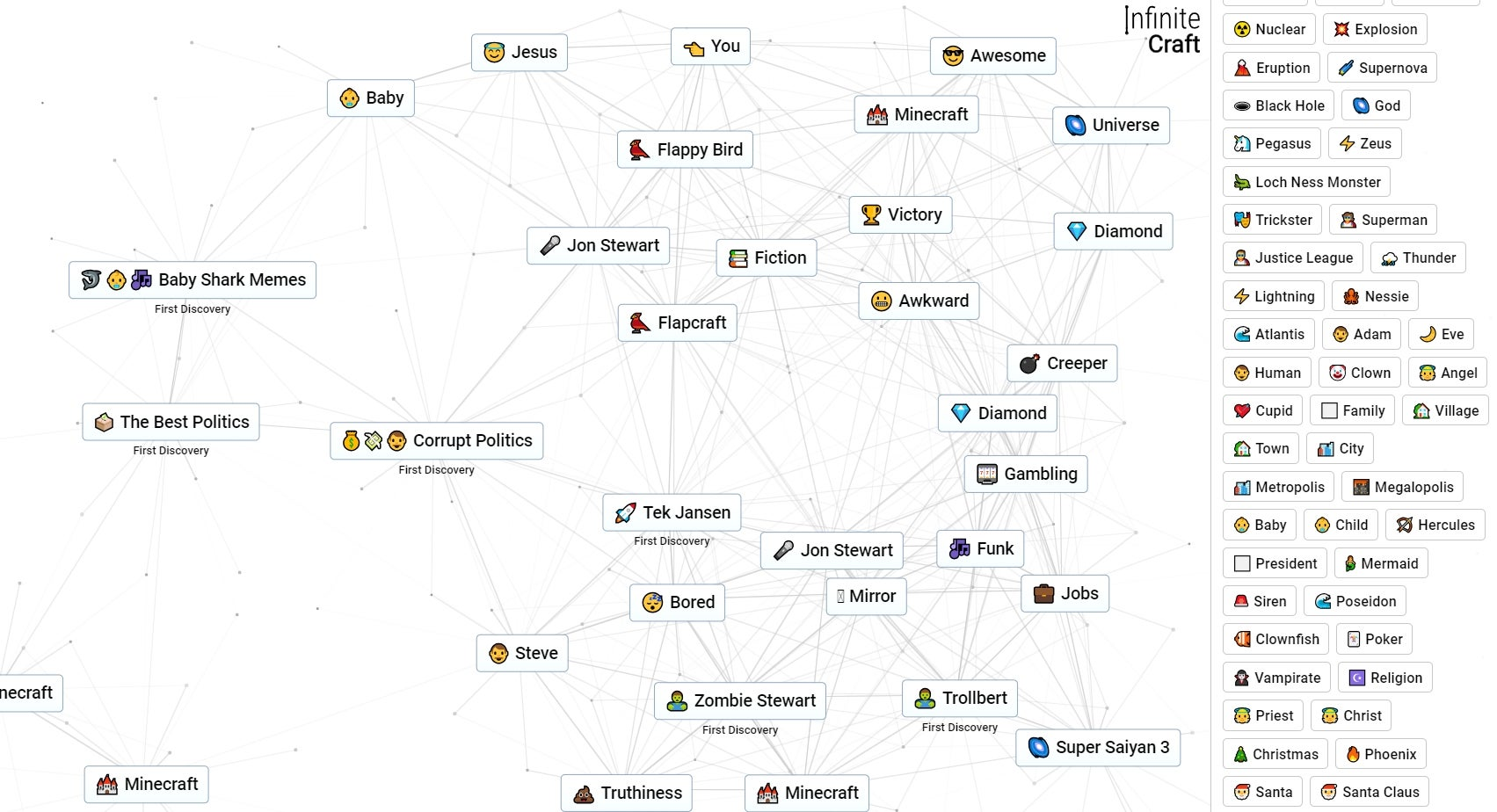}
    \caption{In \emph{Infinite Craft}, the player combines what begins as a simple set of atomic elements into increasingly complex entities, with an LLM dictating the product resulting from arbitrary combinations. Image used with permission.}
    \label{fig:infinite_craft}
\end{figure}

Another strength of LLMs is language synthesis. Huang and Sun \cite{huang2023create} used GPT-3 to generate new words from the combination of two user-selected words. These are used to progress in a text-based game scenario, where the player wins by unlocking goal words. Similarly, language synthesis is leveraged by \emph{Infinite Craft}\footnote{\url{https://neal.fun/infinite-craft/}}, an \enquote{alchemy} game, in which the player combines elements to produce new ones (see Fig.~\ref{fig:infinite_craft}). In \emph{Infinite Craft}, the player begins with a set of core elements (water, fire, wind and earth). But while the former have a set of interactions defined manually by the designer, \emph{Infinite Craft} prompts Llama~2 \cite{touvron2023llama} to imagine the product of the combination of these elements \cite{pcmaster2024infinitecraft}. Judging from gameplay, it appears that for each distinct combination, Llama is prompted to produce the result only once, with the product stored in a database for future reference. Thus seemingly anything in the language model's vocabulary might \enquote{emerge} from the combination of these elements, including all 50 states\footnote{\url{https://x.com/FeralFlex/status/1758332430136615298}}, \enquote{Dream}\footnote{\url{https://x.com/slutzsmp/status/1760394123243135169}}, and the fictional \enquote{Super Stonedosaurus Tacosaurus Rex}\footnote{\url{https://x.com/pvtspicy/status/1759316982984237139}}. On occasion, the model can choose to return one of the combined elements, or refuse to combine (e.g. very lengthy or complex) elements.

\subsection{Automated Designer}\label{sec:designer}

A key role of AI in games \cite{yannakakis2018ai_games} is the algorithmic generation of game content such as levels and visuals, or even entire games. Unlike a Game Master who creates a game via natural language\textemdash meant to exist in the \enquote{theater of the mind} of the players\textemdash the aim of procedural content generation (PCG) is to create content intended for use in a digital game and thus it is required to satisfy certain constraints such as playability and aesthetic quality.

Any PCG method that is trained on available content corpora fits under the Procedural Content Generation via Machine Learning (PCGML) paradigm \cite{summerville2018pcgml}. Strictly speaking the original PCGML framework of 2018 did not consider LLMs; instead it relied on machine learning methods such as autoencoders and LSTMs. However, important challenges of PCGML remain when considering LLMs for PCG: notably, the reliance on high-quality, machine-readable datasets from human-authored levels. While some datasets exist for arcade game levels \cite{summerville2016vglc}, for most games the content remains both unavailable and protected by intellectual property (IP) laws. We revisit this issue in Section \ref{sec:issues_llm_in_games}. Prior work in PCG has demonstrated that tile-based game levels can be reliably generated with sequence-based prediction models (e.g. LSTMs) from a modest set of examples, by treating such levels as linear sequences of tile types in raster order \cite{summerville2016super, zakaria2022procedural}. 

\begin{figure}[t]
    \centering
    \includegraphics[width=0.55\columnwidth]{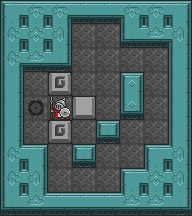}
    \caption{A level for the puzzle game \textit{Sokoban} generated by GPT-3, visualized with the Griddly tileset \cite{bamford2021griddly}. Image used with permission.}
    \label{fig:gpt3_sokoban}
\end{figure}

More recently, this approach has been extended to modern LLMs pre-trained on natural language (instead of sequence models trained from the ground up).
Todd \textit{et al.} \cite{todd2023level} fine-tuned a GPT-2 model on a large dataset of \emph{Sokoban} (Thinking Rabbit, 1982) levels and, at test time, sampled from the model to produce novel puzzles (see Figure~\ref{fig:gpt3_sokoban}). Interestingly, their results indicate that while the GPT-2 model struggles when the fine-tuning dataset is restricted in size, GPT-3 (and, presumably, larger models released since then) are better able to accommodate limited training sets.

Using a similar approach, {MarioGPT} trains a GPT-2 model on a relatively small dataset of \textit{Super Mario Bros} (Nintendo, 1985) levels \cite{sudhakaran2024mariogpt}. MarioGPT overcomes the issue of data scarcity by using the initial dataset as the starting point for an evolutionary algorithm. Existing levels are selected and then sections of the level are mutated by sampling from the GPT model and then correcting the border between the re-generated section and the rest of the level with a similarly-trained BERT (i.e. bi-directional) model \cite{devlin2019bert}. This approach produces a large and diverse set of playable levels, despite starting from less than 20 levels.

The above GPT-based level generation approaches also show the promise of incorporating natural language instructions to produce \textit{conditional} level generators, either by prefixing game levels in the training dataset with desired level characteristics \cite{todd2023level} or by embedding user instructions and allowing the model to attend to the embedding during generation \cite{sudhakaran2024mariogpt}.
A recent example of the latter is \textit{Cardistry} \cite{lyman2024cardistry}, which used GPT-3.5 to transform a short personal narrative into a set of playing cards. Along with text information, the LLM creates prompts for DALL-E \cite{betker2023improving}, completing the generation pipeline with playing card art.

User requests however are typically fuzzy and can be misinterpreted by the LLM. Hu \textit{et al.} \cite{hu2024minecraft} first \enquote{refine} user requests for the generation of \textit{Minecraft} structures using an LLM, adding knowledge specific to the domain (such as block palette and building dimensions). Then, a second LLM is employed to generate the description of the structure that can be interpreted and placed in the game.

While most applications of LLMs as automated designers involve some mapping from the LLM's natural language output to other game-specific formats, LLMs can naturally be applied to games where the content is itself just natural language. In this vein, LLMs have been leveraged for the generation of New York Times \textit{Connections} puzzles~\cite{merino2024making}, which involve having a user group 16 words into 4 groups of 4 words each, with the challenge being for the user to infer what common semantic theme unites each group. 

\begin{figure}[t]
    \centering
    \includegraphics[width=0.98\columnwidth]{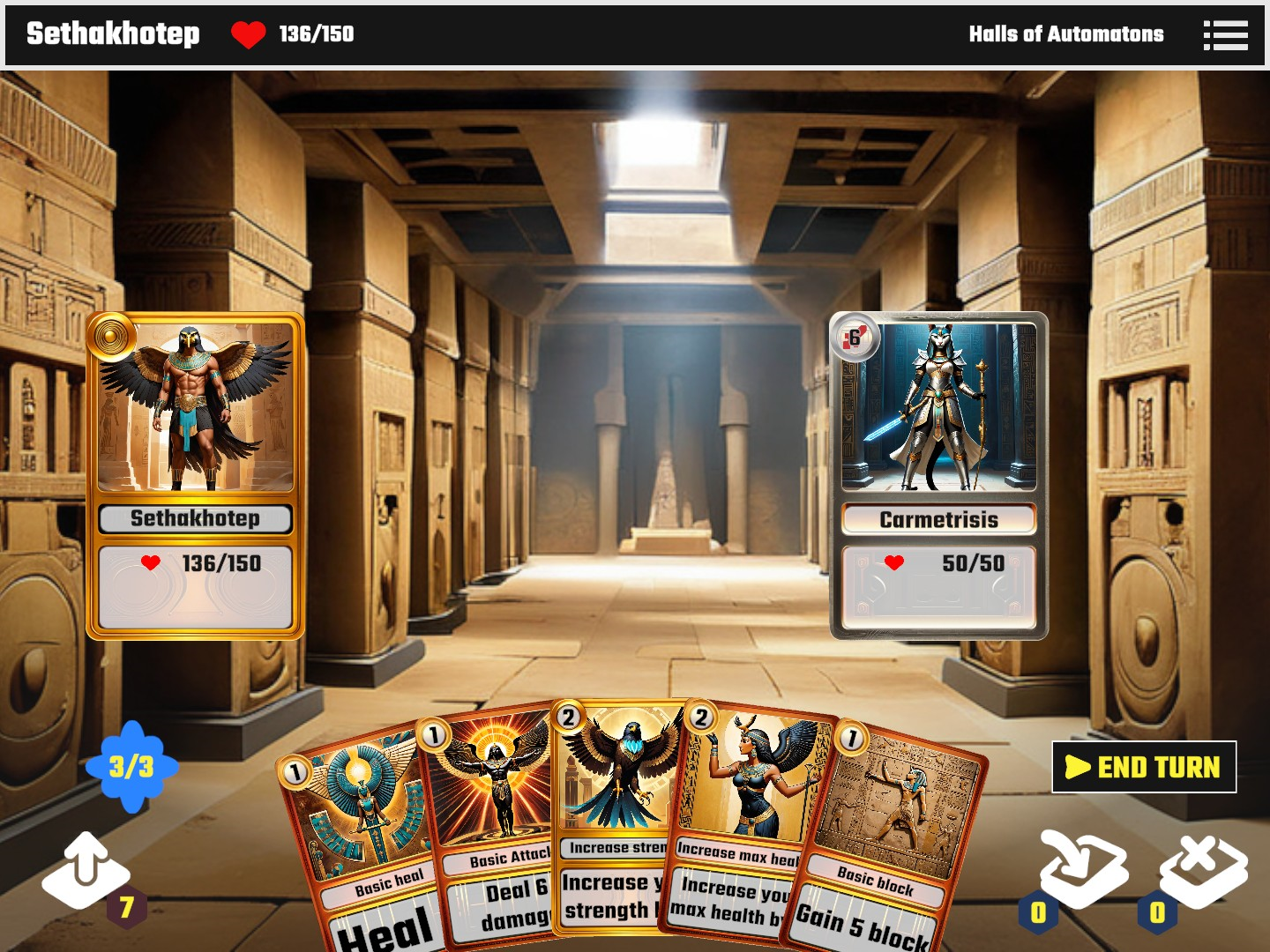}
    \caption{A screenshot of a \textit{CrawLLM} game instance, generated for the theme of Ancient Egypt \cite{zammit2024crawllm}. In \textit{CrawLLM}, the LLM generates themes, stories, characters, and locations for a card-based dungeon crawler game while Stable Diffusion  generates the visuals. Image used with permission.}
    \label{fig:crawllm}
\end{figure}

Unlike the above examples, LLMs can also produce instructions for other LLMs or Foundation Models (FMs) without a user request in the first place.
In CrawLLM \cite{zammit2024crawllm}, the Mixtral 8x7B LLM \cite{jiang2024mixtral} generates the theme, visual style, and even enemy descriptions which act as blueprints to guide additional LLM queries for producing player-facing narrative (e.g. introductory text) or textures and animations via Stable Diffusion \cite{Rombach_2022_StableDiff} for re-theming a dungeon crawler game with card-based combat mechanics. In CrawLLM, the game design (code and card details) are pre-authored by the human, while LLMs and foundation models re-theme the assets to provide a visually and narratively novel experience every time (see Fig.~\ref{fig:crawllm}).

Finally, LLMs can generate new games by writing game code directly. For instance, the {GAVEL} system \cite{todd2024gavel} combines evolution with LLMs to generate board games. In GAVEL, a language model fine-tuned on a dataset of board game programs written in the {Ludii} description language \cite{browne2019ludii} acts as the mutation operator within the evolutionary loop.

\subsection{Design Assistant}\label{sec:design_assistant}

An AI for design assistance can provide several benefits to the creative process. Depending on the type of tool, type of AI, and type of creative process, the AI can minimize development time and cost, reduce human effort, support collaboration among members of a design team, or elicit a user's creativity \cite{liapis2014thesis}. So far, in games, most of the AI-powered design assistant tools focus on autocompleting a human's in-progress design \cite{smith2011tanagra} or providing many possible suggestions for the designer to consider \cite{liapis2013sketchbook,migkotzidis2021susketch,charity_baba_2022,gallotta_preference-learning_2023, charity2023preliminary,torii2023lottery}.
Ideally, we would want an LLM that can act as a human colleague that we can bounce off ideas to and collaborate with. Such LLMs are still beyond the current state of the art \cite{dhamani2023tyranny}. Existing tools implement co-creating \cite{yannakakis2014micc} LLMs at different levels of control, which we can explore under the existing Co-Creative Framework for Interaction Design \cite{rezwana2023designing}.
Focusing on the \textit{interaction} over the artifact, the LLM can be of \textit{conceptual assistance}, providing high-level guidance which is not game-ready. This would require that the designer adapts and curates the AI output in a way that fits their own vision and the constraints of the game. This implies that the LLM contributes very little in the actual artifact generation, merely providing new suggestions. By allowing the LLM to also \textit{refine} or \textit{transform} the artifact, we see a \textit{procedural assistance} by the LLM. Interacting with the designer, the LLM can produce increasingly more final versions of the intended artifact. The LLM is expected to understand the context of the game for which the content is intended, in order to provide meaningful assistance. However, the LLM does not need to produce a final, playable artifact but could instead simply provide the next creative step for discussion with the designer \cite{liapis2013world}. Moreover, the designer is ultimately responsible for curating and adapting the generated content, as well as deciding when the co-creative process is completed \cite{barth2023ghostwriter, guzdial2019interaction,novick1997mixedinitiative}. Finally, if the LLM is allowed to \emph{directly} create and alter the artifact based on user requirements, we say it provides \textit{production assistance}. This is the closest level to PCG (see Section \ref{sec:designer}), but is different in that the designer remains in control and can refine their specification or reject a created artifact (versus an autonomous generator which directly sends content to the player). As expected, however, the AI operates in a much more constrained space in this scenario as it must account for all other game mechanics (the design of which are presumed finalized) and designer goals which are somehow encoded or presumed via learned designer models \cite{liapis2013designermodeling}.

One can argue that existing interfaces with LLMs and Large Multimodal Models (LMMs) act as design assistants. The designer provides their specifications and receives one (in LLMs) or multiple (in AI image generators) suggestions that they can further refine. Many creatives report using such interfaces for brainstorming and concept development \cite{vimpari2023adapt}, including game developers \cite{boucher2023professionals}. However, the applicability of LLMs as design assistants is somewhat limited, reverting only to conceptual assistance. Similarly, their potential for refining an existing idea (i.e. offering procedural assistance) is underexplored, as we discuss in Section \ref{sec:roadmap}.

Conceptual assistance is thus the easiest for LLMs, and is the first case explored in games. Charity \textit{et al.} \cite{charity2023preliminary} envision design assistance as a tool that combines the game description provided by the user with existing knowledge of similar games to suggest possible game features back to the user. The suggested features are fairly generic, few-word guidelines (e.g.: \enquote{learn new combat}) which would need extensive design effort and creativity to transform into an implementable and coherent game design. When asking an LLM for specific game features to implement in a digital game, players found them less compelling than human-designed ones \cite{anjum2024inksplotch}. The suggestions, however, were still useful for game designers, as they provided a different perspective that could kickstart their creation process. A thorough analysis of strengths and weaknesses of LLMs in this role can be drawn from \textit{Project AVA} \cite{peacock2024ava}, a non-commercial digital game developed at Keywords Studios with the assistance of LLMs and LMMs for multiple aspects of the typical game development pipeline. LLMs are shown to help greatly in giving inspiration to the designers (albeit not being creative themselves), provide simple starting code for the game logic, and even assist developers by revealing errors in the code during development. LLMs however often fall short on anything more involved or requiring further domain knowledge, such as requesting code for specific game logic or understanding functionality in UI elements. Similarly, LMMs required a lot of tuning by human artists, but provided a solid foundation for concept art and proof-of-concept user interfaces.

Since production assistance is also close to traditional PCG pipelines, it is also understandably explored for games. Nasir and Togelius \cite{nasir2023practical} used GPT-3 to generate levels for the \textit{Metavoidal} (Yellow Lab Games, 2022) brawler game from a prompt that describes the level's features (e.g. width and height) while a human curates and edits the results to ensure playability. This curated set of levels is then used for further fine-tuning, potentially automating the generative process. Kelly \textit{et al.} \cite{kelly2023there} instead use GPT-4 to generate stories in natural language while abiding to logic constraints, assisting story writers. Instead of generating the final artifact directly, Kumaran \textit{et al.} \cite{kumaran2023endtoend} use the LLM to parse game level parameters from the user request. The natural language input to the LLM contains information about both what the user is looking for and properties that the level of \textit{Future Worlds}---a  game-based learning exhibit about environmental sustainability \cite{rowe2017play}---should posses, such as difficulty, size, type of sustainability problem, and simulation goal metrics. A collection of candidate game levels are generated from a single user request. Each candidate is then evaluated by a RL agent, and the best game is presented to the designer via the Unity game engine.

\begin{figure}[t]
    \centering
    \includegraphics[width=0.98\columnwidth]{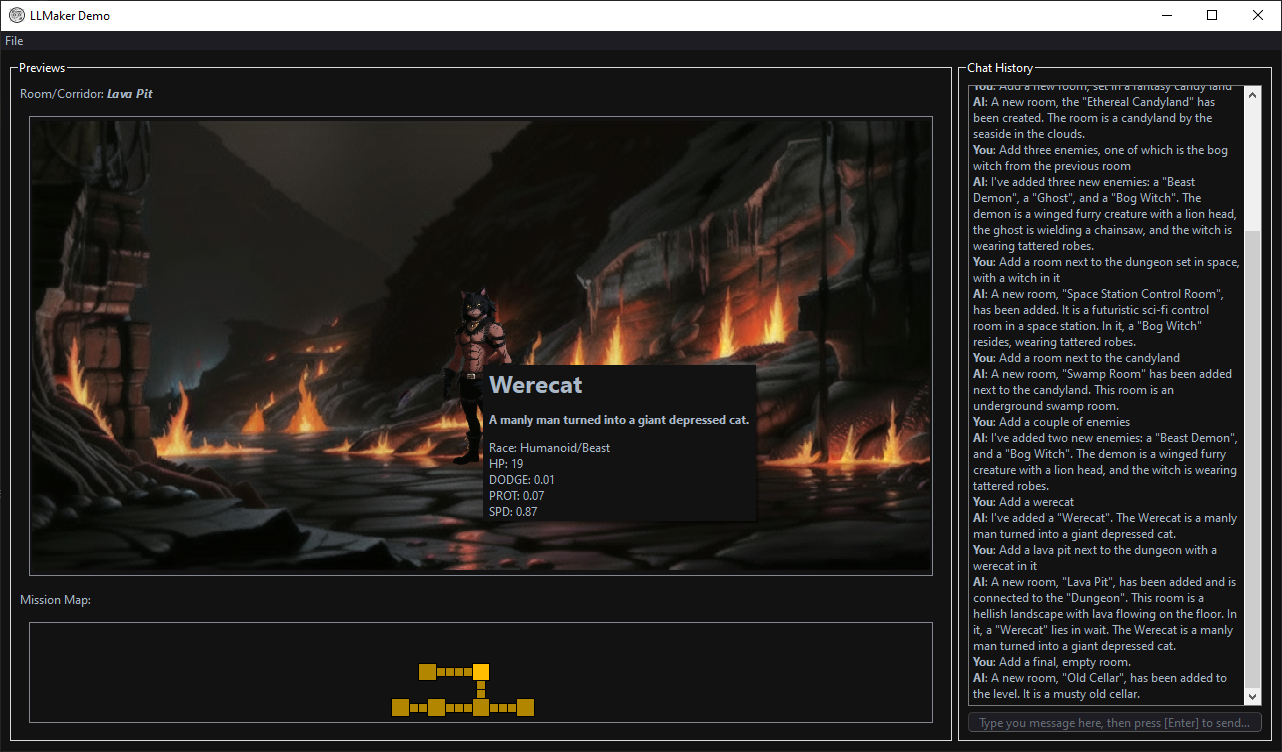}
    \caption{A screenshot of the \textit{LLMaker} digital game content design assistant tool \cite{gallotta2024llmaker}. In \textit{LLMaker}, a human designer makes requests in natural language to alter a dungeon level for a video game, editing both the level layout and the entities within. The LLM satisfies the designer requests by calling the appropriate function, ensuring that the content adheres to domain constraints. Image used with permission.}
    \label{fig:llmaker}
\end{figure}

Examples of procedural assistance are very few: \textit{LLMaker} \cite{gallotta2024llmaker} (see \Cref{fig:llmaker}) is a chat-only level editor where the user can chat with the LLM and request changes to be made to the game level and its content. Natural conversation with the LLM allows the designer to explore the current state of the level (for example, the user may ask \enquote{What are the enemies in the entrance room?}) or draw inspiration for the next changes to make (for example, by asking \enquote{What kind of enemies should I add to this corridor?}). When the user requests a change to be made to the level, the LLM translates their request into a valid function call, ensuring content consistency and adherence to domain constraints \cite{gallotta2024funccall}. Parameters for the function call that have not been specified by the user are generated by the LLM itself, biased by the existing level and overall user preferences. For example, if the user asks \enquote{Change the name of the banshee to Scary Ghost} the LLM will not only change the enemy's name but also their description to reflect the change, even though this was not specified explicitly by the user (see Fig.~\ref{fig:llmaker}) In addition to the level and its content, \textit{LLMaker} employs Stable Diffusion models to generate the graphical assets for the artifacts, based on descriptions generated by the LLM itself. As noted above, so far research has mostly focused on either LLMs for conceptual assistance (putting significant onus on a human designer) or as production assistance (leveraging a human designer as curator). The conversational nature of LLMs, however, seems particularly well-suited for procedural assistance when designing content; we revisit this underexplored area of research in Section \ref{sec:roadmap}.

\section{A Roadmap for Future Applications of LLMs in Games}\label{sec:roadmap}

The previous section attempted to group current research in LLMs and games into a typology focused on the roles an LLM is asked to play. As part of this exercise, we identified a number of roles that have been heavily researched. Unsurprisingly, the role of player and automated designer have received most attention: this matches the general trends within AI and games research more broadly \cite{yannakakis2018ai_games}. Following general trends in Game AI for playing or generating content, LLM-based approaches are likely to flourish via community events, benchmarks and competitions, with first steps already being taken in this direction \cite{taveekitworachai2023chatgpt4pcg}. Based on the roles listed in Section \ref{sec:roles}, we identify below some gaps found in the literature, and lay out possible research directions that leverage the power of LLMs in new ways.

While academic interest in design assistance within games has blossomed in the last decade, we find that the potential of LLMs has so far been underutilized. LLM design assistants either ask too much of a human designer in terms of creative interpretation and actual development \cite{charity2023preliminary} or too little, demoting them to content curator \cite{nasir2023practical}. Past research in mixed-initiative systems \cite{yannakakis2014micc} assumes a more co-creative initiative from both human and machine, and the power of LLMs as conversational agents matches the original vision of a creative dialogue between initiatives \cite{novick1997mixedinitiative}. Therefore, a promising unexplored direction lies in a more procedural assistance (see Section \ref{sec:design_assistant}) where the LLM not only produces output but also reasons about it to the human designer. LLMs seem especially well-suited for this task, as the context is retained and the designer can iteratively refine past products that the LLM has generated. However, concerns of LLMs' limited memory may arise (see Section \ref{sec:limitations}) in long-term design processes. On the other hand, iterative refining is not as straightforward for other state-of-the-art technologies such as LMMs, despite some promising results via e.g. \textit{InstructPix2Pix} \cite{brooks2023instructpix2pix}. It is expected that such applications will raise new challenges in terms of hallucinations, explainability \cite{zhu2018explainable}, capturing or modeling designer intent \cite{liapis2013designermodeling}, and more. We discuss such challenges further in Section \ref{sec:issues_llm_in_games}.

While we identified player assistance as an important role that LLMs can play, we have found little work that targets any aspect of this beyond re-theming designer-defined hints \cite{rist2024using}. The conversational ability of LLMs make them ideally suited for tutorial writing or hint-giving, especially in short snippets as provided e.g. by a conversational agent. However, it is important to note that LLMs often hallucinate or overfit to the corpus they have been trained on, and may be challenged, for instance, to summarize or lookup specific rules given a game manual. Similar limitations were identified when using an LLM as assistant to a human Game Master \cite{Kelly2023TTRPGsupport}, where the LLM could not find information in the pre-written adventure when asked a question about the scene. Other technologies (as simple as a database search query) could be used instead, with the LLM undertaking only the task of converting the found information into a natural language utterance. Beyond mere hint-giving, however, an LLM could also act as a more hands-on player assistant, taking over more trivial tasks (such as managing minutiae of one city in a strategy game). This is also powerful for Game Master assistance, as the LLM can keep track of locations visited and NPCs met, or look up rules. In both cases, addressing the issue of hallucinations and consistency will need to be addressed, which we review in Section \ref{sec:limitations}.

Another role seemingly well-suited for LLMs that has received limited attention is that of commentator or reteller. Work so far has focused on automating the commentary of streamers or eSport casters \cite{renella2023towards}.  While this direction is still largely uncharted, there are more directions that could leverage LLMs for streamer assistance rather than automation (and replacement). Rather than narrate events occurring within the game (or video stream), LLMs can summarize the audience interactions and engagement levels \textemdash thus acting as a commentator not of the game but of the audience watching it. This could allow a human streamer to better keep track of topics discussed in the chat, and engage as needed without having to read every comment. While this has been identified as a research direction for AI already \cite{liapis2023audiences}, it has yet to be implemented. Under the role of streamer assistance, issues of explainability of the LLM's commentary would become pertinent (e.g. to address one audience member by name); we revisit this in Section \ref{sec:limitations}. It is worth noting that streamers have already begun to explore AI assistance to their streams. To the best of our knowledge, the most relevant example is YouTube user Criken who plays alongside an AI assistant as a conversational agent, having a dialogue with an otherwise out-of-the-box LLM either embedded within the game\footnote{\url{https://youtu.be/dQ-7-r5aM1U}} or as part of the stream\footnote{\url{https://youtu.be/KhE9NhUqtBc}}. This example is not explicitly targeted for reacting to in-game events or audience interaction (and acts more as NPC than commentator) but indicates that some streamers are open to the use of such technology for their craft.

Despite a few attempts to leverage LLMs for games user research (see Section \ref{sec:analyst}), there is much unexplored potential in this direction. LLMs so far are used to cluster gameplay logs \cite{wang2024player2vec,rasajski2024behave}, but they could also explain the groupings in the form of e.g. play personas \cite{canossa2009playpersona,canossa2009tombraider} described in natural language. Such a task would raise issues of explainability and privacy more broadly (see Section \ref{sec:limitations}), and would likely still involve a game designer or user researcher in the loop for quality assurance. More importantly, moving from these log-based clusters towards capturing the player's experience or emotional state \cite{yannakakis2023affective} remains an open challenge. 
In principle, an LLM could predict affective state transitions such as ``the game is more engaging now'' and thereby adapt the game environment to elicit a supposedly more engaging experience for the player. Learning such transitions builds on the experience-driven procedural content generation paradigm \cite{yannakakis2011experience} but with an LLM acting as the player experience model. Future research could explore how LLMs can be fine-tuned to represent and infer player experience transitions based on in-game observations and demonstrations of experience. Two challenges need to be addressed for this: (a) representations of game states as natural language and (b) hallucinations of human experience. For the former challenge, potentially leveraging work of LLM players (see Section \ref{sec:player}) and how they pass the game state to the LLM seems a promising first step. For the latter challenge, however, current LLMs struggle to capture user intent during conversation \textemdash let alone more ill-defined concepts such as players' emotion or engagement \cite{pinitas2023massive}. 
Current datasets on affect in games are formatted as continuous or categorical variables, often fluctuating over time \cite{melhart2022again}, which would be challenging to format as text without processing. While perhaps using language as input or output for the player model requires some innovative pre-processing or more advanced LLM technologies, the underlying GPT architecture shows promise already. Broekens \emph{et al.} showed that ChatGPT could detect emotion in English text \cite{broekens_2023_finegrained}, although admittedly games include many more modalities (e.g. visuals, audio) than pure text, which is mainly relegated to narrative \cite{liapis2021chapter}. We expect more research on player modeling powered by transformers, if not LLMs directly, such as leveraging behavior transformers \cite{shafiullah2022behavior} to imitate human playtraces grouped by playstyle \cite{pfau2023dungeon}.

To wrap up, we believe that every role an LLM could be called to play in (or around) a game identified in Section \ref{sec:roles} could benefit from additional attention. This technology remains nascent and changes are forthcoming which may address several limitations we identified above and more extensively in Section \ref{sec:limitations}. The natural language capabilities (especially for text generation) make LLMs ideal conversational assistants (for a player, a designer, a GM, or a streamer). The ability of LLMs to consume and reason from text corpora also opens new possibilities for automated design moving beyond tile-based level generation (which needs carefully crafted corpora) and more towards open-ended content such as game narratives \cite{johnson2023theme, kreminski_2019_cozy, fernandez_2012_procedural, dormans_2010_adventures,alnassar2023questville,kumaran2023scenecraft} or even game design documents. The potential of LLMs in that regard is already voiced by many evangelists in the field, but research on actual implementation of such ideas and on addressing the IP concerns they may raise (see Section \ref{sec:issues_llm_in_games}) are still forthcoming.

While the focus of this paper is on what LLMs can do for games, we do not underestimate what games can do for LLMs. One of the watershed moments for AI and games research was the article by Laird and Van Lent naming games as the ``killer app'' for human-level AI \cite{laird2001killerapp}. This remains true for LLMs today: games are ideally poised for LLM research. Not only do games produce rich multimodal data (ideal for e.g. LMMs), but there also exist rich corpora of text and multimodal data produced by players, viewers, fans, etc. Game text data, such as transcripts, have already been used to train LLM players \cite{meta2022human,wang2023voyager}. On the other hand, LLMs struggle with both spatial reasoning and planning by their very nature, while most games rely heavily on both aspects. From strategy board games and digital games (where long-term planning is crucial) to first-person shooters (which hinge on precision in spatial reasoning and a reactive plan for reaching the enemy base), such games remain state-of-the-art testbeds for gameplaying AI \cite{alphastar,milani2022minerl} and will likely be fraught arenas for LLM research. Games also hinge on long-term interactions, especially in the case of LLM-based GMs (see Section \ref{sec:gamemaster}). Games can thus form testbeds or benchmarks to explore the limits of recollection under different context lengths, a critical limitation of LLMs detailed in Section \ref{sec:limitations}. In terms of game design tasks, we also note that games are complex constrained problems, with hard constraints on e.g. levels that can be completed \cite{smith2012constrainable}, but also soft constraints regarding game balance between competing players in multi-player games \cite{karavolos2017patterns,pfau2023dungeon}, or the progression and pacing of a single-player experience \cite{smith2011tanagra}. While some LLMs can handle some hard constraints via, say, function calling \cite{gallotta2024funccall}, this may not be possible for more complex or more constrained game domains. Moreover, soft constraints would need to be conveyed to the LLM in more nuanced ways. Game benchmarks specific to LLMs have already started to emerge \cite{taveekitworachai2023chatgpt4pcg}, but identifying critical game-based challenges for LLMs, appropriate and interesting benchmarks, and (ethically sourced) data for training or fine-tuning LLMs remains an open question.

\section{Limitations of LLMs in games}\label{sec:limitations}

Large language models have exciting potential for games, but they also come with inherent limitations. Mainly, LLMs suffer from hallucinations \cite{duan_2023_shifting,manakul_2023_selfcheckgpt}, meaning that they will output plausible but false statements simply because they are a probable sequence of words. Hallucinations are inevitable, given how the world is described to the machine \cite{strasser_2023_pitfalls}; LLMs lack grounding, so the text they generate is detached from constraints of reality. Yet LLMs always \enquote{act} confidently in their responses, even when wholly mistaken. Indeed, Hicks \textit{et al.} argue that the term \textit{AI hallucinations }misrepresents how ``the models are in an important way indifferent to the truth of their outputs'' \cite{hicks2024bullshit}. LLMs  are shown to also output responses that are wrong even though the LLM has access to information that proves otherwise \cite{bian_2023_influence,karpinska_2023_large,gekhman_2023_trueteacher}. In the context of digital games, these limitations affect certain applications of LLMs more than others, for example NPCs may hallucinate quests that do not exist in the game, or a player assistant may provide suggestions to the user based on wrong assumptions.

Another limitation is that LLMs sometimes struggle to capture user intent. This is especially evident with expressions of sarcasm \cite{zhou_2023_evaluation}. The ability to capture user intent is important for applications of LLMs that converse directly with the player. Many LLMs misunderstand user requests \cite{liu_2023_summary}, and clarifying to the LLM multiple times leads to frustration. This limitation is most relevant to cases where the LLM is in direct conversation with the user, e.g. as design assistant, player assistant, or Game Master. Depending on how much the LLM output controls the user experience (e.g. as Game Master or offering production assistance to a human designer), the inability to capture user intent can be a frustrating experience.

On a larger scale, LLMs suffer from losing context, and struggle with continuity. This is because the \enquote{memory} of an LLM is constrained by its context size, which limits the extent of its inputs and outputs, as well as its response time due to the attention mechanism \cite{vaswani2017attention}. The longer the conversation, the less likely it is that the LLM will recall early events \cite{li2023-memory}. In digital games, it is possible to separately summarize the game events (see \Cref{sec:reteller}) and process them as part of the input to the LLM. As a game progresses past a few game sessions, however, this summary may still be too long, or details of increasing significance will be omitted, thus leading to a degraded performance. This is especially relevant for roles requiring long-term engagement, such as LLM-powered retellers or Game Masters. In \emph{Infinite Craft} (see Section \ref{sec:mechanic}), this is handled by an external database that stores and looks up past combination rules, ensuring consistency in future uses of the same mechanic. However, LLMs could theoretically tackle this issue directly.

Recent models have tried to address this recollection issue by increasing the context length, with some of the larger models encompassing 128K or even 10M tokens \cite{geminiteam2024gemini15}. Despite this being adequate for a wide range of applications, it may still fall short when applied to long-term tracking of game states. In particular, massive multiplayer online games offer a simulation space with a large intricate domain of actions and interactions, which scales exponentially with the number of agents (players or otherwise) participating.
Researchers have also tried to address the context limit by including compressive memory into the attention mechanism of the LLM \cite{munkhdalai2024infinite}, in an attempt to create a seemingly infinite context length. The authors of \cite{munkhdalai2024infinite}, however, acknowledge its current limitation, partly due to the difficulty of selecting and compressing the data which should be \enquote{memorized}. A different approach proposed by Fountas \textit{et al.} \cite{fountas2024humanlike} draws inspiration from cognitive science to equip LLMs with episodic memory, greatly reducing context length limitations during information retrieval.

A Retrieval-Augmented Generation (RAG) system \cite{Lewis2020RAG} could address this limitation, drawing from a database containing vector representations or other latent representations of pertinent text or data. When the text generator processes a sequence, the RAG system would retrieve similar entries from this external data source. This would hypothetically provide a streamlined archive of game events and actions for the LLMs to consult in order to generate a consistent narrative progression. 

Another challenge is that currently LLMs are trained to be highly compliant to the users' requests. For an LLM assistant, this is not a cause for concern, but in the role of a Game Master this can create issues. Human GMs frequently curb the more exotic player requests which could drastically diverge from the game narrative or which would result in an unrecoverable disruption of a required sequence of game events. An LLM Game Master would try to accommodate for even the most bizarre requests, with little consideration for the consequential impacts to any predetermined game events.

Yet another limitation of LLMs that prevents their application in mainstream media is their cost. Running AAA games and LLMs in parallel on consumer hardware is infeasible \cite{valve_steam_survey} due to their computation requirements. If one wants to integrate LLMs in games, they would have to host the models on their own servers or access existing models via APIs. Additionally, the cost of querying LLMs is a recurring cost, and cannot be properly estimated beforehand. This kind of problem is also affected by the scale of LLMs-powered games or tools. Similarly to how server costs increase with the number of active players in massive multiplayer online games, the more players use a LLM over multiple play sessions, the more the game developers or publishers will have to bear the financial burden. The monetary cost of this approach can be prohibitive or difficult to estimate for real-world applications. The game need not even be played by other players: to evaluate the performance of their simulations with multiple LLM-based NPCs, Park \textit{et al.} ran simulations for several days with a cost of ``thousands of dollars in token credits'' \cite{park2023simulacra}. While promising techniques to reduce the costs of running LLMs exist \cite{yue2024large,yu2024affordable}, these are not yet widespread and require further engineering to set them up properly.

Perhaps due to the above limitations, the implementation and deployment of LLMs in digital game applications is still very limited. A digital game is a domain where responsiveness is vital for players, so it follows that LLMs should also be able to provide their responses quickly. Unfortunately, while research on more efficient and faster architectures is being carried out \cite{miao_2023_efficient}, the real-time application of LLMs is still not plausible. This is especially evident in other domains such as design applications, where \enquote{real time} responses are generated in around 30 seconds to over a minute \cite{delatorre_2023_llmr}.

\section{Ethical Issues with LLMs in games}\label{sec:issues_llm_in_games}

With the improvement of AI methods applied to games over the recent years, many questions regarding their ethics and real-world impact have been raised \cite{melhart2023ethics}. Using LLMs raises ethical issues regarding sustainability, copyright, explainability, privacy, and biases. Naturally, each of these issues has serious implications in the field of games.

The reliance of LLMs on training data and training time raises concerns regarding their carbon footprint. Beyond training costs, inference over the model's lifespan has a greater environmental impact due to constant querying \cite{chien_reducing_2023, khowaja_chatgpt_2023}. Factors like renewable and local energy, better model architectures, and more meaningful (and thus less wasteful) training data can mitigate this. In the context of LLMs for digital games, sustainability remains crucial, considering the carbon footprint of frequent queries during gameplay (e.g. for Game Master or NPC responses, or for LLM-powered players). This is especially pertinent if the LLM is intended to run locally, on consumer-level hardware which are usually powered by non-renewable sources.

When it comes to copyright, issues apply to the input data, the output data, and the model itself. LLMs trained on data under copyright is an unfortunate common practice \cite{github_copilot_lawsuit}, deservedly raising public outrage \cite{rothchild_2022_copyright,github_copilot_hackernews}. 
The models themselves have different copyright licenses applied, which can also lead to artifacts they generate to fall under the public domain \cite{jiang2023mistral,zhang2024tinyllama}. For the game industry, matters of IP and copyright are extremely important. This is as much a concern regarding having the company's copyrighted content somehow used as training data by competitors, as it is about LLMs producing material that the company cannot copyright. It is important to note here that, at least when it comes to the latter concern, the role the LLM takes is very pertinent. If an LLM or LMM produces content automatically (see Section \ref{sec:designer}), past legal consensus in the USA indicates that the material can not be copyrighted \cite{copyright2022entrance}. If an LLM or LMM acts as an \enquote{assistive tool} \cite{copyright2023zarya} to a designer (especially for conceptual assistance, see Section \ref{sec:design_assistant}) then the extensive and impactful human effort needed to transform these concepts into game design and game art likely makes the final product eligible for copyright \cite{copyright2023zarya}. The limited rulings in copyright courts regarding this, however, and the \enquote{likely} caveat we include in our own text, understandably would make game companies hesitant to tread in untested waters for major game IPs beyond e.g. small-scale indie productions \cite{peacock2024ava,sun2023language}. For researchers, however, the ethical issues of copyright breach and exploitation by large corporations, and the public outcry for the above, leave a bad taste and make research in LLMs less palatable \cite{lamb2023storm}.

In applications, understanding how a final result or product is reached is extremely crucial, particularly when a product is iteratively refined as with design assistants (see Section \ref{sec:design_assistant}). This is a problem of explainability \cite{zhu2018explainable}, whereas LLMs are inherently opaque in their generation process. Liu \textit{et al.} \cite{liu_trustworthy_2023} highlight different methods to improve the explainability of language models, such as concept-based explanations or saliency maps. Particularly for LLMs, the self-explanation applied via the chain-of-thought \cite{kojima_2022_large} reasoning has received attention by the research community \cite{chen_2024_boosting,mondal_2024_kamcot}. While this method adds a layer of explained reasoning to the generated output, there are multiple examples in the literature that demonstrate how this reasoning may just be an illusion of reasoning capabilities. Such examples include disregarding the provided reasoning in the final output \cite{turpin_language_2023}, or reaching the correct solution via incorrect steps in math problems \cite{frieder_mathematical_2023}. In the domain of games, explainability is paramount across roles, ensuring gameplay coherence and user engagement.

Replicability of an application's behavior is equally crucial. When applying LLMs to digital games (especially as game mechanics, NPCs, or automated designers) one would expect their output quality to not change over time. This is not the case for closed-source LLMs: even when using the same model name, the same request at one time can generate content that is vastly different from a past iteration \cite{karkaj2023prompt}. In this case, developers may have to consider switching to open-source models with open weights, such as those hosted on the popular HuggingFace Transformers library \cite{wolf2020transformers}. An additional benefit of switching to self-hosted models, model size notwithstanding, is the additional level of privacy that is guaranteed to the users. Querying local models ensures all messages remain within the application, whereas interacting with models hosted via APIs entails that conversations are exchanged over third-party websites. A developer might be willing to share conversation logs with an API provider\footnote{Such as ChatGPT, which shares conversations by default unless opted out} for model improvement. However, users may not be aware of this practice or its implications. Local deployment of LLMs has been democratized by making models more accessible on lower-end hardware, relying on the widespread adoption of the GGUF format\footnote{Details of this file format are available at \url{https://github.com/ggerganov/ggml/blob/master/docs/gguf.md}} and the release of different versions of the same model at varying degrees of quantization \cite{jacob2018quantization}. The quantization of an LLM usually results in a loss of performance, but this is usually considered a valid trade-off for the reduced model size to load on VRAM. 
Combined with friendly APIs for running LLMs locally, such as Open WebUI\footnote{\url{https://openwebui.com/}} and LM Studio\footnote{\url{https://lmstudio.ai/}}, it is possible to run LLMs in a more controlled fashion. The more pertinent technological breakthroughs lie in compacting size and carbon footprint while retaining high-quality LLM outputs.

Finally, biases emerge as LLMs are trained on a large corpus, usually scraped from the (Western-focused part of the) internet. This allows models to capture a current reality snapshot, which is advantageous for a conversational or question-answering model, though it requires curating this data from different kinds of biases. Some biases, such as social stereotypes, could be targeted and alleviated; others, such as exclusionary norms, pose greater challenges. In games, we identify two main concerns when interacting with an LLM: toxic behavior, and stereotypes or incorrect notions. Toxic behavior is a harmful property that a language model may learn from its training corpus, which often contains text from community-based fora or social platforms. Tools that combat toxic language in digital games are constantly evolving, with some even blocking chat messages before they are delivered to the user \cite{jia_2022_ingame,patent:20210370188}. Therefore, similar applications could theoretically be developed to target toxic outputs from language models. Unlike human players, however, when an LLM plays the role of an NPC, it should align with the game themes and avoid any kind of toxic language or racial slurs. This requires developers to ensure proper behavior of the model through data cleaning, if the model is trained from scratch, or supplying tailored data if finetuning it to their needs. Addressing \emph{prejudices} such as stereotypes and incorrect notions is complex, as they are not necessarily related to single words or expressions, but instead present themselves as a collection of ideals that can be wrong at best, and harmful at worst. An NPC LLM may exhibit real world stereotypes that can negatively impact the player experience, although we argue that the impact of prejudices from an LLM commentator or Game Master is much stronger and disturbing due to their perceived authority.

\section{Conclusions}

As discussed in this paper, LLMs can take up many different roles that can improve the experience of players in digital games, or enhance the ability of game designers to bring their ideas to life. However, we also highlighted many different challenges specific to the applications of LLMs and intrinsic to the nature of LLMs and the ecosystem that surrounds them. Despite technical, ethical, and legal challenges posed by LLMs, it is not realistic to ignore the impact that this research will likely have on both Game AI research and the game industry. We expect to see many new technical innovations from LLM researchers and corporations. Anticipating this, we propose promising directions where LLMs could be applied to games in the future.

\section*{Acknowledgments}
This work has been supported by the European Union’s Horizon 2020 research and innovation programme from the AI4media project (Grant Agreement No. 951911), and by the US National Science Foundation under the Graduate Research Fellowship Program.


\newpage

\begin{IEEEbiography}[{\includegraphics[width=1in,height=1.25in,clip,keepaspectratio]{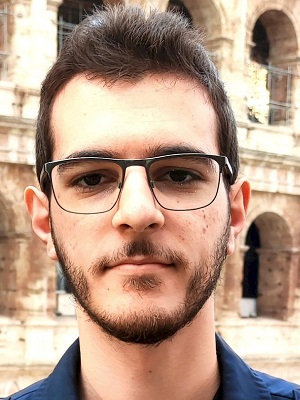}}]
{Roberto Gallotta} is a second-year {Ph.D.} student in {Game Research} at the Institute of Digital Games, University of Malta, researching dynamic evolutionary computation and mixed-initiative co-creation for the aide of video game designers. Prior to joining the Institute of Digital Games, he was a junior researcher at Araya Inc., Tokyo, for a year. He has published multiple papers, mostly focused on procedural content generation and evolutionary computation with games as a domain, and co-organized the \enquote{ALife for and from video games} workshop at ALIFE2023.
\end{IEEEbiography}

\begin{IEEEbiography}[{\includegraphics[width=1in,height=1.25in,clip,keepaspectratio]{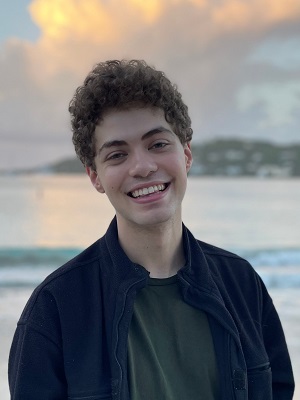}}]
{Graham Todd} is a fourth-year PhD student in {Computer Science} in the Game Innovation Lab at NYU Tandon. His research focuses on the intersection of language games, and the ways in which people and algorithms can generate novel games and goals. He is particularly interested in what games can teach us about their players. He has published a number of papers on topics ranging from modeling the process of game generation with evolutionary algorithms to interrogating the ways that LLMs understand language through word games.
\end{IEEEbiography}

\begin{IEEEbiography}
[{\includegraphics[width=1in,height=1.25in,clip,keepaspectratio]{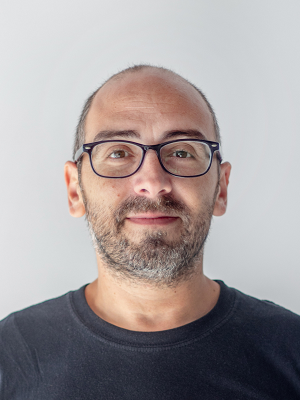}}]
{Marvin Zammit} is a researcher at the Institute of Digital Games, University of Malta, where he is currently reading for a {Ph.D.} in {Game Research}. His research revolves around computational creativity, procedural content generation, evolutionary computation, quality diversity, and educational games. His primary interest lies in the application of machine learning algorithms in building practical pipelines for game development. He is also an experienced developer of games and interactive installations.
\end{IEEEbiography}

\begin{IEEEbiography}
[{\includegraphics[width=1in,height=1.25in,clip,keepaspectratio]{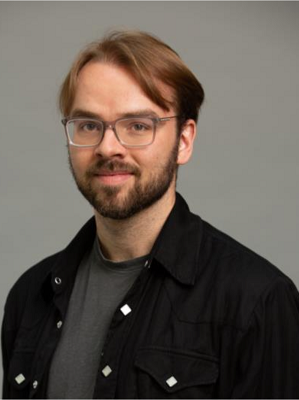}}]
{Sam Earle} is a fifth-year PhD student in {Computer Science} in the Game Innovation Lab at NYU Tandon. His research focuses on open-ended learning in terms of the automatic generation of diverse playable game environments, and the training of robust embodied agents within these games. He has also investigated using foundation models to enable text-guided environment generation, with an eye toward steering ever-complexifying environment- and agent-generation loops toward human-relevant content. 
\end{IEEEbiography}

\begin{IEEEbiography}
[{\includegraphics[width=1in,height=1.25in,clip,keepaspectratio]{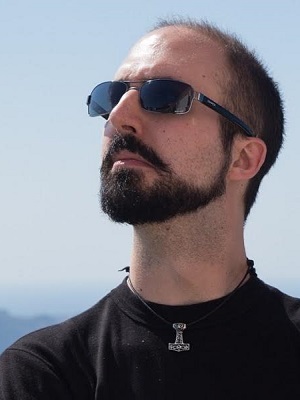}}]
{Antonios Liapis} is an Associate Professor at the Institute of Digital Games, University of Malta, where he bridges the gap between game technology and game design in courses focusing on human-computer creativity, digital prototyping and game development. He received the Ph.D. degree in Information Technology from the IT University of Copenhagen in 2014. His research focuses on Artificial Intelligence in games, human-computer interaction, computational creativity, and user modeling. He has published over 140 papers in the aforementioned fields, and has received several awards for his research contributions and reviewing effort. He serves as Associate Editor for the {IEEE Transactions on Games}, and has served as general chair in four international conferences, as guest editor in five special issues in international journals, and has co-organized 15 workshops. 
\end{IEEEbiography}

\begin{IEEEbiography}
[{\includegraphics[width=1in,height=1.25in,clip,keepaspectratio]{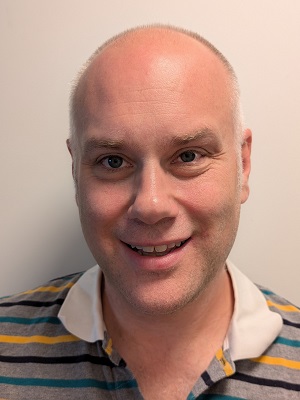}}]
{Julian Togelius} is an Associate Professor in the Department of Computer Science and Engineering at New York University, and director of the NYU Game Innovation Lab. He is also a co-founder and research directo of the game AI company modl.ai. Julian's research focuses on games for AI and AI for games; for example, procedural content generation via reinforcement learning, open-ended learning in generative environments, and LLM-guided game creation. Julian got his PhD in {Computer Science} in 2007 from the University of Essex in England, and he has also worked in Switzerland and Denmark despite being Swedish. He was Editor-in-Chief of IEEE Transactions on Games from 2018 to 2021.
\end{IEEEbiography}

\begin{IEEEbiography}[{\includegraphics[width=1in,height=1.25in,clip,keepaspectratio]{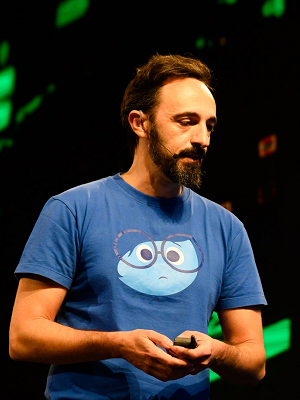}}]
{Georgios N. Yannakakis} (F'24) is a Professor at the Institute of Digital Games, University of Malta (UM) and a co-founder of modl.ai. He received the PhD degree in Informatics from the University of Edinburgh in 2006. Prior to joining the Institute of Digital Games, UM, in 2012 he was an Associate Professor at the Center for Computer Games Research at the IT University of Copenhagen. He does research at the crossroads of artificial intelligence, computational creativity, affective computing, advanced game technology, and human-computer interaction. He has published more than 350 papers in the aforementioned fields and his work has been cited broadly. His research has been supported by numerous national and European grants (including a Marie Skłodowska-Curie Fellowship) and has appeared in \emph{Science Magazine} and \emph{New Scientist} among other venues. He is currently the Editor-in-Chief of the {IEEE Transactions on Games} and an Associate Editor of the {IEEE Transactions on Evolutionary Computation}. Georgios is an IEEE Fellow.
\end{IEEEbiography}

\end{document}